\documentclass[sigconf]{acmart}

\usepackage{xcolor}
\usepackage{colortbl}
\usepackage{multirow}

\AtBeginDocument{%
  }

\copyrightyear{2024}
\acmYear{2024}
\setcopyright{acmlicensed}\acmConference[GLSVLSI '24]{Great Lakes Symposium on VLSI 2024}{June 12--14, 2024}{Clearwater, FL, USA}
\acmBooktitle{Great Lakes Symposium on VLSI 2024 (GLSVLSI '24), June 12--14, 2024, Clearwater, FL, USA}
\acmDOI{10.1145/3649476.3658784}
\acmISBN{979-8-4007-0605-9/24/06}

\begin{document}

\title{GeckOpt: LLM System Efficiency via Intent-Based Tool Selection}


\author{Michael Fore}
\affiliation{%
  \institution{Microsoft Corporation}
  \city{Reston}
  \state{VA}
  \country{USA}
}
\email{mifore@microsoft.com}

\author{Simranjit Singh}
\affiliation{%
  \institution{Microsoft Corporation}
  \city{Silicon Valley Campus}
  \state{CA}
  \country{USA}
}
\email{simsingh@microsoft.com}

\author{Dimitrios Stamoulis}
\affiliation{%
  \institution{Microsoft Corporation}
  \city{Redmond}
  \state{WA}
  \country{USA}
}
\email{stamoulis.dimitrios@microsoft.com}


\begin{abstract}
In this preliminary study, we investigate a GPT-driven intent-based reasoning approach to streamline tool selection for large language models (LLMs) aimed at system efficiency. By identifying the intent behind user prompts at runtime, we narrow down the API toolset required for task execution, reducing token consumption by up to 24.6\%. Early results on a real-world, massively parallel Copilot platform with over 100 GPT-4-Turbo nodes show cost reductions and potential towards improving LLM-based system efficiency.
\end{abstract}



\begin{teaserfigure}
  \includegraphics[width=\textwidth]{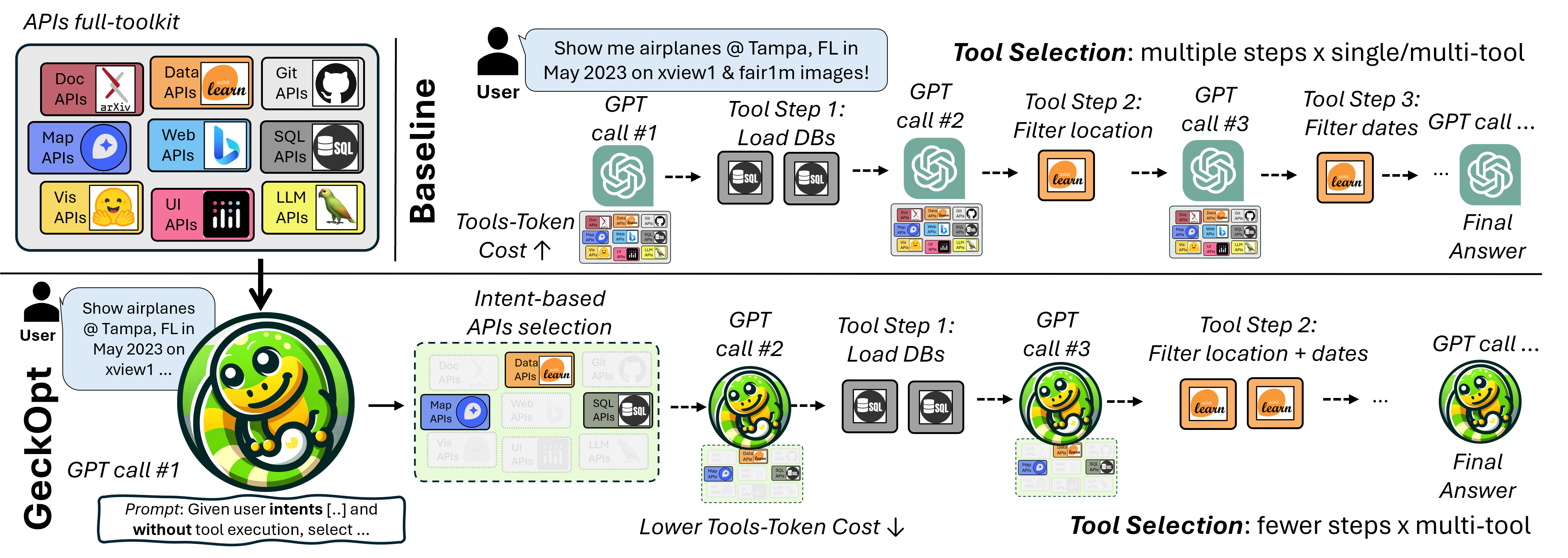}
  \caption{Intent-based tool selection with \texttt{GeckOpt}. For each user query, the agent identifies relevant API libraries - without initially executing specific tools! Preliminary results indicate that streamlining the toolset reduces token costs associated with function calls by approximately 25\%, while encouraging multi-tool execution over fewer GPT requests.}
  \Description{An overview of our proposed GeckOpt framework.}
  \label{fig:teaser}
\end{teaserfigure}


\maketitle

\begin{table*}
  \caption{Query categorization based on \textit{intent} following~\cite{singh2024geoengine}.}
  \label{tab:intents}
  \begin{tabular}{ccl}
    \toprule
    Intent Example & Related Query & APIs \\
    \midrule
    Load$\rightarrow$Filter$\rightarrow$Plot & Plot xview1 images around Tampa Bay, FL, USA  & \texttt{SQL\_apis()}, \texttt{data\_apis()}, \texttt{map\_apis()}  \\
    UI/Web Navigation & Search Bing for ``System-efficient LLM prompting''? & \texttt{web\_apis()}, \texttt{UI\_apis()} \\
    Information Seeking & Which model to use for airplane detection? & \texttt{wiki\_apis()} \\
    \bottomrule
\end{tabular}
\end{table*}

\begin{table*}
  \caption{\texttt{GeckOpt} intent-based tool-selection; enhanced agent performance on the \texttt{GeoLLM-Engine-10k} baselines~\cite{singh2024geoengine}.}
  \label{tab:results}
  \begin{tabular}{lccccccc}
    \toprule
    \textit{GPT-4 Turbo (0125)} & Correct. Rate $\uparrow$ & Success Rate $\uparrow$  & Obj. Det F1 $\uparrow$  & LCC R $\uparrow$ & VQA  Rouge-L $\uparrow$  & Tokens/Task $\downarrow$ \\
    \midrule
    CoT Zero-Shot~\cite{singh2024geoengine} & 80.88\% & 77.35\% & 87.99\% & 96.56\% & 65.29\% & 23.6k\\    
    \rowcolor{teal!15} CoT Zero-Shot  + \texttt{GeckOpt} & 79.13\% & 77.03\% & 86.03\% & 96.46\% & 63.21\% & \textbf{18.48k}\\
    CoT Few-Shot~\cite{singh2024geoengine} & 84.01\% & 80.00\% & 88.40\% & \textbf{99.89\%} & 67.65\% & 25.8k\\
    \rowcolor{teal!15} CoT Few-Shot + \texttt{GeckOpt} & 83.11\% & 79.26\% & 88.19\% & 98.53\% & 68.66\% & \textbf{19.45k}\\
    \midrule
    ReAct Zero-Shot~\cite{singh2024geoengine} & 84.27\% & 80.03\% & \textbf{89.34\%} & 98.83\% & 68.11\% & 26.7k\\
    \rowcolor{teal!15} ReAct Zero-Shot + \texttt{GeckOpt}  & 83.87\% & 79.46\% & 88.02\% & 98.74\% & 68.73\% & \textbf{20.38k}\\
    ReAct Few-Shot~\cite{singh2024geoengine} & \textbf{84.31\%} & \textbf{81.11\%} & 83.85\% & 99.63\% & \textbf{69.37\%} & 32.5k\\ 
    \rowcolor{teal!15} ReAct Few-Shot + \texttt{GeckOpt}  & 84.10\% & 80.17\% & 83.31\% & 99.01\% & 69.26\% & \textbf{25.14k}\\    
    \bottomrule
\end{tabular}
\end{table*}

\section{Preliminary Investigation}

Large language models (LLMs) augment platforms by facilitating the reasoning, planning, and execution of system tools via natural language instructions, thanks to advancements in prompting techniques~\cite{lu2023chameleon, wei2023chainofthought, qin2023toolllm}. These techniques, acting as LLM-based planners, determine the sequences of tools (\textit{i.e.}, underlying system APIs) to complete tasks. Despite the impressive LLM performance, this usually comes at considerable hardware cost.

In this initial examination, we embrace these advancements, while adopting a system-oriented view to identify bottlenecks in tool-calling processes. By analyzing an in-house large-scale Copilot system, namely \texttt{GeoLLM-Engine}~\cite{singh2024geoengine, singh2024geoqa, jian2023stable}, we observe that current compositional tool-calling methods often result in \textit{multi-step} $\times$ \textit{single-tool} executions (Fig.~\ref{fig:teaser} top). This contrasts with the more desirable \textit{multi-step} $\times$ \textit{multi-tool} approach, which aggregates multiple API calls within a single GPT step (Fig.~\ref{fig:teaser} bottom). Since each step translates to a GPT request and token consumption, the total number of steps significantly impacts system cost.

Our objective is to strike a balance between the inclination of compositional reasoning - breaking down tasks into multiple steps for task comprehension for the LLM-planner - and minimizing the system steps required for task completion. Our early analysis reveals that, as is to be intuitively expected, presenting the LLM with a narrower selection of tools, combined with a clearer understanding of the task, encourages the aggregation of more tools per step. To this end, we draw inspiration from the concept of \textit{LLM intent}, where state-of-the-art methodologies demonstrate LLMs' capability to discern ``user intent'' and utilize it to categorize tasks into granular user-defined templates~\cite{zhou2023webarena}.

\texttt{GeckOpt}: Our methodology begins with an offline phase where tasks are mapped to intents and associated tools with minimal human involvement (Table ~\ref{tab:intents}). Next, \textbf{at runtime} and for each prompt, we first engage the LLM agent to determine the task's intent and to identify the relevant subsets of API libraries - ahead of pinpointing specific tools at this stage! This intent-driven ``gating,'' while it incurs the minor cost of an extra API call, effectively narrows down the tool selection pool for all subsequent compositional prompting to proceed as usual. The result is a reduction in token requirements and a move towards executing multiple tools within a single API call, optimizing both token usage and system resource allocation. Given that our approach is \textit{fully} GPT-driven, it adeptly handles failures where the initial API selection is incorrect, with the agent being instructed via prompting to revert to the full toolset.

\section{Preliminary Results}

\textit{Experimental setup}: as a representative Copilot platform, we use our \texttt{GeoLLM-Engine} framework~\cite{singh2024geoengine} that spans several remote sensing tasks, open-source datasets, and API tools for loading, filtering, processing, and visualizing data. We report agent performance metrics on the \texttt{GeoLLM-Engine-5k} benchmark for CoT~\cite{wei2023chainofthought} and ReAct~\cite{yao2023react} prompting. We refer the reader to~\cite{singh2024geoengine} for further details.

\textit{Results}: Table~\ref{tab:results} shows that \texttt{GeckOpt} reduces token usage across various baselines up to 24.6\% with small performance variation. While we note slight deviations, possibly attributable to employing non-zero temperature settings in LLM function calling to foster creative answers in visual question answering, we observe that cost reductions come at negligible performance degradation within 1\% in terms of success rates. Extrapolated across numerous user sessions over our cloud-native platform, such token reduction by almost a quarter can culminate in substantial cloud cost savings. 

\textit{Limitations - Future Work}: We note that this is a preliminary investigation where geospatial tasks are well-suited to intent-based granularity. An important next step is to confirm the generalizability of the methodology across various function-calling benchmarks. Last, as our study considers only cloud endpoints, we are currently expanding to local LLM execution. Given the correlation between token usage and operational costs, we anticipate additional improvements for LLM system utilization, underscoring the potential of intent-based tool selection for enhancing hardware efficiency.

\bibliographystyle{ACM-Reference-Format}
\bibliography{main}

\end{document}